\documentclass{article} % For LaTeX2e
\usepackage{iclr2027_conference,times}

\usepackage{amsmath,amsfonts,bm}

\def\eqref#1{equation~\ref{#1}}
\def\1{\bm{1}}

\DeclareMathAlphabet{\mathsfit}{\encodingdefault}{\sfdefault}{m}{sl}
\SetMathAlphabet{\mathsfit}{bold}{\encodingdefault}{\sfdefault}{bx}{n}

\usepackage{hyperref}
\hypersetup{hidelinks}
\usepackage{url}
\usepackage{soul}
\usepackage[table]{xcolor}
\usepackage{graphicx}
\usepackage{booktabs}
\usepackage{subcaption}
\usepackage{placeins}
\usepackage{wrapfig}
\usepackage[capitalize]{cleveref}
\usepackage{multirow}
\usepackage{fontawesome5}

\graphicspath{{images/publication/}{images/}}
\renewcommand{\cite}{\citep}

\newcommand{\method}{\textsc{MemBodied}}

\title{\method: Recurrent Associative Memory for Vision-Language-Action Models}

\author{Tej Deep Pala\textsuperscript{1}\thanks{Corresponding author:
\href{mailto:tejdeep001@e.ntu.edu.sg}{\texttt{tejdeep001@e.ntu.edu.sg}}
} \quad
Navonil Majumder\textsuperscript{1} \quad
Bryce Goh\textsuperscript{2} \quad
Raphael Yee\textsuperscript{2} \\
\textbf{Jianfei Yang}\textsuperscript{1} \quad
\textbf{Liming Chen}\textsuperscript{3} \quad
\textbf{Soujanya Poria}\textsuperscript{1} \\
\textsuperscript{1}Nanyang Technological University \quad
\textsuperscript{2}Griffin Labs \quad
\textsuperscript{3}\'Ecole Centrale de Lyon \\\\
\faGithub~Code: \href{https://github.com/declare-lab/MemBodied}{\texttt{https://github.com/declare-lab/MemBodied}} \\
\faGlobe~Project Website: \href{https://declare-lab.github.io/MemBodied}{\texttt{https://declare-lab.github.io/MemBodied}}
}

\iclrfinalcopy
\begin{document}

\maketitle

\begin{abstract}

Vision-Language-Action models provide a strong foundation for general-purpose robot control, yet a vast majority of policies do not preserve and leverage episode-level information beyond current observation. This limitation is consequential in history-dependent manipulation tasks that depend on information available only in past observations. Retaining past observations in context can aid in recovering this information, but at the significant cost of ever-growing bloated context and inference latency. We thus introduce \method, a fixed-size episodic memory with two complementary components: an associative state that records interactions across policy calls and an episode anchor that preserves a compact representation of the initial scene as a reference. At each policy call, the model conditions action generation on the current input and the memory components, rather than directly using past observations. Across five evaluated RMBench tasks requiring memory, \method\ achieves $7.81\times$ the mean success rate of a stateless policy and $2.98\times$ of vanilla recurrent memory, while outperforming the strongest memory-augmented baseline by $1.3\times$ with $10\times$ fewer added parameters. On the fully observable LIBERO-Long suite, it reached 90.6\%, a 5.4\% improvement over the stateless $\pi_0$ policy. These findings support \method\ as a practical alternative to expanding the policy context for history-dependent manipulation.

\end{abstract}

\section{Introduction}

Vision-language-action (VLA) models are increasingly a mainstay for general robot control as they map semantic representations learned from large vision-language datasets to continuous or token-based actions. RT-2 and OpenVLA formulate actions within vision-language backbones \cite{brohan2023rt,pmlr-v270-kim25c}; Octo~\cite{team2024octo} learns a diffusion-based generalist policy from heterogeneous robot data; and $\pi_0$ couples a pretrained vision-language model to a continuous flow-matching action expert \citep{black2024pi_0}. More recent models, such as OpenVLA-OFT~\cite{kim2025fine} and $\pi_{0.5}$~\cite{intelligence2025pi_}, have improved control efficiency, action generation, and open-world generalisation. Together, these developments have made VLA policies increasingly capable at complex manipulation tasks.

However, many existing VLA policies condition each policy call on the current observation and the language instruction to predict future action chunks. These approaches discard past information, limiting the policy's ability to complete memory-dependent manipulation tasks. Although action chunking promotes consistency within a predicted horizon, it does not preserve observational information across subsequent policy calls. For example, if a robot moves an object away from its initial position and must later return the object, the current observation may no longer reveal the correct destination. This is a temporal state aliasing problem, where identical or similar current observations may require different actions depending on the past observations.

Although preserving history is desirable, placing past observations directly into the policy context is costly. Retaining the complete history causes the context length, memory consumption, and inference cost to grow throughout an episode. A fixed observation window bounds these costs but may discard task-relevant evidence once it falls outside the window. Moreover, the relevant evidence may be a short event separated from the current decision by many intermediate observations. An effective memory must therefore preserve relevant information without growing in size or inference cost over the episode.

Recent work addresses this problem by compressing historical observations~\cite{koo2026hamlet,wang2026nativemem}, maintaining recurrent latent representations~\cite{cherepanov2026mu,li2026remem,guo2026chameleon}, or explicitly storing and retrieving historical information~\cite{fang2025sam2act,shi2026memoryvla,li2025map,haresh2026self}. Methods that retain compressed observations as history tokens can still enlarge the policy context as new information is added. Recurrent-token methods keep the recurrent state fixed in size, but rely on carried embeddings to implicitly preserve and expose relevant information through repeated transformations. Memory-bank, retrieval, and language-based approaches may instead introduce auxiliary objectives or additional processing pipelines. These considerations motivate a memory that is updated online, remains fixed in size, and retrieves information according to the policy's current state. Associative memory meets these requirements through an explicit read/write structure in which the delta rule specifies how the state is updated while the policy learns what to write, retain, and retrieve using only its action objective.

In this paper, we introduce \method, an episodic memory for VLA control designed to preserve both interaction history and initial-scene information. Its associative memory stores interactions in layer-wise matrices through learned read and write interfaces. After each policy call, the model combines the executed action with its observed visual consequence and writes this interaction using a gated delta update. Because repeated updates may overwrite fine-grained details from early in the episode, the anchor pathway provides persistent access to a compact representation of the initial scene. At subsequent calls, information retrieved from both memory mechanisms conditions the action network. This division of roles allows past interactions and initial-scene evidence to inform action prediction while keeping the memory footprint and access cost independent of episode length.

We address three research questions in this paper: \textbf{RQ1:} How effective is \method\ for memory-dependent manipulation compared with existing approaches? \textbf{RQ2:} How computationally efficient is \method\ during inference? \textbf{RQ3:} Does \method\ preserve performance on fully observable or Markovian manipulation where memory is not required?

Across five memory-dependent RMBench tasks, \method\ achieves $7.81\times$ the mean success rate of a stateless $\pi_0$ policy and $2.98\times$ that of vanilla recurrent memory, outperforming NativeMEM's compressed-history approach by $1.30\times$. Evaluated on a second VLA backbone ($\pi_{0.5}$), \method\ yields a $3.87\times$ performance increase over the stateless policy. On physical robot experiments across three tasks, \method\ demonstrates an $8.0\times$ improvement in success rate. Crucially, \method\ achieves these gains while reducing inference latency by $91.9\%$ relative to NativeMEM. On LIBERO, \method\ achieved 95.1\% mean success rate, comparable to $\pi_0$'s result of 94.2\%. Notably, on the long-horizon LIBERO-Long suite, \method\ reached 90.6\%, exceeding the $\pi_0$ result of 85.2\% by 5.4 percentage points. These results demonstrate that \method\ improves history-dependent control while remaining competitive on fully observable manipulation tasks.

\section{Related Work}
\label{sec:related_works}

\paragraph{Memory representations for manipulation.} HAMLET~\cite{koo2026hamlet} and NativeMEM~\cite{wang2026nativemem} compress observations into historical tokens. $\mu$VLA~\cite{cherepanov2026mu}, ReMem-VLA~\cite{li2026remem}, and AVA-VLA~\cite{xiao2026ava} carry recurrent latent states across control steps. SAM2Act~\cite{fang2025sam2act}, MemoryVLA~\cite{shi2026memoryvla}, MAP-VLA~\cite{li2025map}, and Notes-to-Self~\cite{haresh2026self} use memory banks, retrieval, or language-based records. These approaches trade off explicit access to past evidence against the cost of retaining history or introducing additional memory components. 
\method\ instead represents episode history through an evolving associative state and an initial-scene anchor, whose readouts condition the action expert. It therefore neither appends a growing sequence of observations nor retrieves discrete past observations. We discuss these and other related work in more detail in Appendix~\ref{app:extended_related_work}.

\section{Methodology}

\begin{figure}[t]
    \centering
    \includegraphics[width=0.8\linewidth]{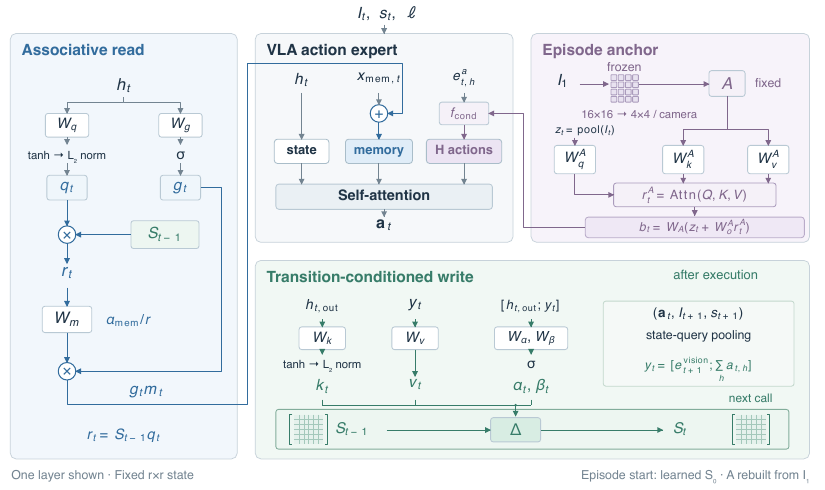}
    \caption{{\method\ architecture.} An associative state records interactions, while the episode anchor preserves a reference to the initial scene.}
    \label{fig:membodied_method}
\end{figure}

\method\ comprises two complementary memory pathways: a recurrent associative state and a fixed visual reference to the initial scene. We first describe the associative update and memory-token readout, followed by the anchor pathway, read/write schedule, and sequence training. We also evaluate attention-steering and hierarchical memory variants for comparison.

\subsection{Recurrent VLA formulation}

We treat manipulation as a history-dependent sequential decision problem where information from prior observations might be necessary to complete the task. We denote the complete episodic memory by $\mathcal{E}_t=(\mathcal{M}_t, A)$, where $\mathcal{M}_t$ is the recurrent associative state and $A$ is the episode anchor. At interaction step $t$, the policy receives current camera observations $I_t$, robot state $s_t$, and language instruction $\ell$ to predict an action chunk $\mathbf{a}_t$ up to horizon $H$:
$$
\mathbf{a}_t=[a_{t,1},\ldots,a_{t,H}]
\sim p_\theta(\cdot\mid I_t,s_t,\ell,\mathcal{E}_{t-1}).
$$
The associative state $\mathcal{M}_{t-1}$ persists across policy calls and is reset to a learned initial state at the start of each episode. The anchor $A$ is constructed from the episode's first observation and remains unchanged until the episode ends.

\subsection{Associative-memory mechanism}

The memory state comprises one associative matrix $S_t^{(l)}$ for each of the $L$ action-network layers. For batch size $B$ and memory rank $r$,
$$
\mathcal{M}_t=\{S_t^{(l)}\}_{l=1}^{L},
\qquad
S_t^{(l)}\in\mathbb{R}^{B\times r\times r}.
$$
In our experiments, we used $r=128$. The state size remains fixed through the episode, i.e., writing new information in memory updates the existing associative matrices rather than appending observations to a growing history.

\subsubsection{Write-value construction}

For an interaction event, we construct the value to be written from the action chunk $\mathbf{a}_t$ and its corresponding observed visual consequence, post-execution, in $I_{t+1}$. 
For each camera $c$, a query derived from the projected robot state $s_{t+1}$ attends to the corresponding visual patch tokens $X_{t+1}^{(c)}$:
$$
e_{t+1}^{(c)}=\operatorname{softmax}\left(
\frac{(W_{pq}\phi_s(s_{t+1}))(W_{pk}X_{t+1}^{(c)})^\top}{\sqrt{D}}
\right)W_{pv}X_{t+1}^{(c)} 
$$
The camera-specific observation encodings $e_{t+1}^{(c)}$ are pooled independently and then averaged:
$$
e_{t+1}^{\mathrm{vision}}
=\frac{1}{C}\sum_{c=1}^{C}e_{t+1}^{(c)}.
$$
Note that, the gradients from the memory-value path are stopped at the visual patch tokens. The pooled next-observation encoding is concatenated with the aggregate of its preceding causal action chunk:
$$
y_t=\left[e_{t+1}^{\mathrm{vision}}\ ;\ \sum_{h=1}^{H}a_{t,h}\right],
\qquad
v_t^{(l)}=W_v^{(l)}y_t.
$$

The resulting representation $y_t$ combines the action summary and observed visual consequence, and each layer-specific projection $v_t^{(l)}$ determines the value written to that layer's associative state. This choice allows the memory value to represent a dynamic interactive transition rather than an isolated observation.

\subsubsection{Gated delta-rule write}

To construct the write key at layer $l$, we transform the post-layer state representation $h_{t,\mathrm{out}}^{(l)}$:
$$
k_t^{(l)}=\operatorname{L_2Norm}
\left(\tanh(W_k^{(l)}h_{t,\mathrm{out}}^{(l)})\right).
$$
The write and retention gate values are then computed from the layer output and write value:
$$
\beta_t^{(l)}
=\sigma(W_\beta^{(l)}[h_{t,\mathrm{out}}^{(l)};y_t]),
\qquad
\alpha_t^{(l)}
=\sigma(W_\alpha^{(l)}[h_{t,\mathrm{out}}^{(l)};y_t]).
$$

The associative key-value state is then updated using a gated delta rule~\cite{yang2025gated}
$$
S_t^{(l)}=
\operatorname{Diag}(\alpha_t^{(l)})S_{t-1}^{(l)}
+\operatorname{Diag}(\beta_t^{(l)})
\left(v_t^{(l)}-
\operatorname{Diag}(\alpha_t^{(l)})S_{t-1}^{(l)}k_t^{(l)}\right)
(k_t^{(l)})^\top.
$$
The first term retains the existing state, while the second term applies a delta correction by comparing the new value with the value already associated with the write key. The retention gate $\alpha_t^{(l)}$ controls how much of the existing state remains, while the write gate $\beta_t^{(l)}$ controls the strength of the correction. The update could therefore revise an existing association without indiscriminately accumulating every value.

\subsection{Associative retrieval and memory variants}

At each layer $l$, the relevant memory readout $r_t^{(l)}$ is retrieved from the respective associative matrix $S_{t-1}^{(l)}$ using a query $q_t^{(l)}$ constructed from the state token representation entering the layer, $h_{t,\mathrm{in}}^{(l)}$:
$$
q_t^{(l)}=\operatorname{L_2Norm}
\left(\tanh(W_q^{(l)}h_{t,\mathrm{in}}^{(l)})\right),
\qquad
r_t^{(l)}=S_{t-1}^{(l)}q_t^{(l)}.
$$

We compare the following three variants of conditioning the action network on readout $r_t^{(l)}$:
\begin{enumerate}
    \item \textbf{Attention steering (\method{}-AS):} Projections of the associative readout produce additive corrections to the attention query and output representations of the action network.
    \item \textbf{Hierarchical memory (\method{}-H):} An LSTM-like recurrent cell is updated from the associative matrices controlled by a learned forget gate and added back, augmenting the base gated-delta update.
    \item \textbf{Memory-token readout (\method):} The associative readout is projected to the width of the action network and added to a dedicated contextual token before self-attention.
\end{enumerate}

\subsubsection{\method: memory-token injection}

In \method, the action-network suffix is arranged as
$$
[\text{state token},\ \text{memory token},\ \text{$H$ noisy action tokens}].
$$
The memory token is a learned vector that is inserted into the suffix at every policy call. The memory token does not persist across policy calls. Instead, the associative matrices carry the recurrent state. At layer $l$, the retrieved vector is linearly transformed and scaled by $\alpha_{\mathrm{mem}}/r$, where $\alpha_{\mathrm{mem}}$ is the memory scaling factor and $r$ is the memory rank. A scalar gate derived from the current state representation then controls its contribution:
$$
x_{\mathrm{mem},t}^{(l)}\leftarrow
x_{\mathrm{mem},t}^{(l)}+g_t^{(l)}m_t^{(l)} \text{, where} 
\quad
m_t^{(l)}=\frac{\alpha_{\mathrm{mem}}}{r}W_m^{(l)}r_t^{(l)}, 
% \text{ and }
\quad
g_t^{(l)}=\sigma(W_g^{(l)}h_{t,\mathrm{in}}^{(l)})
$$

The modified memory token, $x_{\mathrm{mem},t}^{(l)}$,  then participates in self-attention, allowing the action tokens to use the retrieved content. Unlike attention steering, this approach makes the retrieved vector available as contextual content rather than using it only to alter existing attention computations.

\subsection{Fixed episode-anchor memory}

The associative matrices record how an episode evolves, but repeated delta-rule updates can overwrite fine-grained details of the initial scene. We therefore use a separate anchor pathway to preserve a compact, fixed visual reference derived from the episode's first observation. To construct this anchor, we take the frozen vision tokens used in the policy prefix, average-pool each camera's $16\times16$ patch grid into a $4\times4$ grid, and concatenate the pooled tokens across cameras. The resulting anchor $A\in\mathbb{R}^{N_A\times D}$ provides access to the initial scene without retaining the raw observation.

At policy call $t$, the current robot state $s_t$ attends to each camera's visual tokens $X_t^{(c)}$ using the pooling operation defined in the write-value construction. We average the camera-specific outputs to obtain $z_t$, which queries the anchor through rank-$r_A$ cross-attention:

$$
c_t=z_t+W_o^A\left[
\operatorname{softmax}\left(
\frac{(W_q^A z_t)(W_k^A A)^\top}{\sqrt{r_A}}
\right)W_v^A A
\right].
$$

$c_t$ is then projected to the action-network width and repeated across the action horizon:
\vspace{-1pt}
$$
b_t=W_Ac_t,
\qquad
b_{t,h}=b_t,\quad h=1,\ldots,H.
$$

Let $e^a_{t,h}$ denote the action representation at horizon position $h$. The anchor-conditioned representation is computed as
\vspace{-1pt}
$$
\widetilde{e}^{\,a}_{t,h}
=f_{\mathrm{cond}}\!\left(
\left[e^a_{t,h};\,b_{t,h}\right]
\right),
\qquad h=1,\ldots,H,
$$

where $f_{\mathrm{cond}}$ denotes the action network's input-conditioning function. For our experiments, we used $r_A=64$. The anchor remained fixed within an episode and reset at the start of a new episode.

The anchor does not append the first image or its complete token sequence to the policy prefix. Instead, the current observation selectively retrieves a compact spatial reference through cross-attention, leaving the prefix length, token positions, and attention-mask structure unchanged.

\subsection{Causal coordination of reading and writing}

At step $t$, the policy first reads $\mathcal{M}_{t-1}$ and generates $\mathbf{a}_t$. After the action chunk is executed, the environment supplies the resultant visual observation, $I_{t+1}$. The model then associates the hidden state and action chunk from step $t$ with $I_{t+1}$ and writes the association to $\mathcal{M}_t$:

$$
\mathcal{M}_{t-1}
\xrightarrow{\text{read with }(I_t,s_t,\ell)}
\mathbf{a}_t
\xrightarrow{\text{execute}}
I_{t+1}
\xrightarrow{\text{write}}
\mathcal{M}_t.
$$

This delayed-write schedule allows the stored value to include the observed consequence of an action without exposing that future observation to the action that caused it. During inference, the policy caches the preceding layer states and the sampled action chunk from the final denoising step, and performs the write when the next observation arrives. Thus, training and inference use the same causal interpretation of an interaction event.

\subsection{Sequence training}

The delayed write makes sequence-level training necessary. An action loss at a later policy call must propagate through memory operations performed at earlier calls to train the memory parameters. Each training sample comprises a sequence of $N$ observation-action pairs. Consecutive pairs are separated by one action horizon of $H$ environment steps, approximating successive policy calls during execution. Image encoding and policy-input construction are parallelised across the batch and sequence dimensions, while the memory state is propagated sequentially through the sequence. Gradients pass through the full sequence, allowing later action losses to optimise earlier memory operations. The memory parameters are thus learned jointly through the policy's native action objective, without a separate memory-prediction loss.

\section{Experimental Setup}

We evaluate \method\ on memory-dependent manipulation using RMBench~\cite{chen2026rmbench} with the $\pi_0$ and $\pi_{0.5}$ backbones, and on three real-robot tasks. We also use LIBERO~\cite{liu2023libero} to assess whether \method\ preserves performance on fully observable manipulation tasks.

\subsection{Benchmarks and Evaluation Protocol}

RMBench~\cite{chen2026rmbench} is a bimanual simulation benchmark that organises tasks by memory complexity. Its $M(1)$-category of tasks require retention of task-relevant observations from a single past event, whereas the $M(n)$ tasks require information from multiple past events. Our five-task evaluation covered three $M(1)$ tasks: \texttt{put\_back\_block}, \texttt{rearrange\_blocks}, and \texttt{swap\_blocks}; and two $M(n)$ tasks: \texttt{battery\_try} and \texttt{block\_ranking\_try}. We conducted 50 rollouts per task and report each task's success rate and the mean across the five tasks.

As a complementary general manipulation evaluation, we use the four standard LIBERO suites: LIBERO-Spatial, LIBERO-Object, LIBERO-Goal, and LIBERO-Long. Each suite contains ten tasks, and we conduct 50 rollouts per task, yielding 500 rollouts per suite and 2,000 rollouts per checkpoint. We report the mean success rate for each suite and the unweighted mean across all four suites. On both benchmarks, memory is reset at the start of each rollout and persists only for the duration of that episode.

\subsection{Baselines}

Our RMBench-centered comparison includes $\pi_0$-based baselines, published benchmark results, and variants of our method. The locally trained policies share the same $\pi_0$ backbone, LoRA adaptation settings, and number of optimiser steps. These controls align the adaptation recipe and training-step budget, but do not imply identical architectures or compute costs. The full-fine-tuning LIBERO comparisons are reported separately.

\textbf{$\pi_0$-Stateless} takes only the current observation, robot state, and instruction as input, essentially a memory-free baseline. \textbf{$\pi_0$-FrameStack} additionally receives a sequence of past observations and serves as an in-context history baseline. \textbf{$\pi_0$-Hint} augments the stateless policy with simulator-derived textual hints on task progression at inference time, without any training on those hints; it is therefore a privileged-information reference. \textbf{$\pi_0$-Vanilla Recurrent Memory} adapts the associative memory mechanism of Delta-Mem~\cite{lei2026delta}, while \textbf{$\pi_0$-$\mu\text{-VLA}$} adapts the recurrent memory-token mechanism of $\mu$VLA~\cite{cherepanov2026mu}.

We report the Diffusion Policy, ACT, $\pi_{0.5}$, and X-VLA results from \citet{chen2026rmbench}, and use the $\pi_{0.5}$ scores as a full-fine-tuned $\pi_{0.5}$ baseline. With reference to our method, we compare the attention-steering (\method{}-AS), hierarchical (\method{}-H), and memory-token (\method{}) variants of the associative mechanism. Finally, for comparison with NativeMEM~\cite{wang2026nativemem}, we integrate its video-history encoding into our setup and evaluate it both independently and in combination with \method. The full implementation and evaluation details are provided in Appendix~\ref{app:implementation_details} and \ref{app:real_world}.

\section{Results}
\label{sec:results}

\subsection{Performance on memory-dependent manipulation}

\begin{table}[!htbp]
    \centering
    \caption{Success rates (\%) on the five evaluated RMBench tasks. Published policy values are taken from \citet{chen2026rmbench}. The bold figures indicate the best result for each task/column.}
    \label{tab:rmbench_results}
    \footnotesize
    \setlength{\tabcolsep}{3.6pt}
    \renewcommand{\arraystretch}{1.08}
    \resizebox{0.7\linewidth}{!}{%
    \begin{tabular}{@{}lrrrrrr@{}}
        \toprule
        \multirow{2}{*}{Model} & \multirow{2}{*}{\shortstack{Put back\\block}} & \multirow{2}{*}{\shortstack{Rearrange\\blocks}} & \multirow{2}{*}{\shortstack{Swap\\blocks}} & \multirow{2}{*}{\shortstack{Battery\\try}} & \multirow{2}{*}{\shortstack{Block\\ranking}} & \multirow{2}{*}{\textit{Mean}} \\
        & & & & & & \\
        \midrule
        \multicolumn{7}{@{}l}{\emph{Published policies~\citep{chen2026rmbench}}} \\
        Diffusion Policy & 0.0 & 0.0 & 11.0 & 10.0 & 10.0 & 6.2 \\
        ACT & 0.0 & 29.0 & 2.0 & 19.0 & 0.0 & 10.0 \\
        X-VLA & 18.0 & 13.0 & 16.0 & 26.0 & 1.0 & 14.8 \\
        $\pi_{0.5}$ & 11.0 & 13.0 & 16.0 & 16.0 & 6.0 & 12.4 \\
        \midrule
        \multicolumn{7}{@{}l}{\emph{Baselines in our setup ($\pi_0$ backbone)}} \\
        $\pi_0$-Stateless & 6.0 & 2.0 & 8.0 & 16.0 & 0.0 & 6.4 \\
        $\pi_0$-Hint & 10.0 & 0.0 & 0.0 & 0.0 & 0.0 & 2.0 \\
        $\pi_0$-$\mu\text{-VLA}$ & 0.0 & 0.0 & 0.0 & 0.0 & 0.0 & 0.0 \\
        $\pi_0$-FrameStack & 14.0 & 30.0 & 0.0 & 4.0 & 26.0 & 14.8 \\
        $\pi_0$-Vanilla Recurrent Memory & 18.0 & 22.0 & 14.0 & 22.0 & 8.0 & 16.8 \\
        \midrule
        \method-AS & 32.0 & 26.0 & 12.0 & 20.0 & 14.0 & 20.8 \\
        \method-H & 16.0 & 50.0 & 14.0 & 28.0 & 8.0 & 23.2 \\
        \method\ w/o anchor memory & 34.0 & 76.0 & 16.0 & 30.0 & \textbf{32.0} & 37.6 \\
        \method & \textbf{40.0} & \textbf{92.0} & \textbf{56.0} & \textbf{40.0} & 22.0 & \textbf{50.0} \\
        \bottomrule
    \end{tabular}
    }
\end{table}

\paragraph{Recurrent memory and its formulation both affect performance.} To assess the effectiveness of \method\ for memory-dependent manipulation (RQ1), we compare it with baselines and \method\ variants across five RMBench tasks. As shown in \cref{tab:rmbench_results}, \method\ reaches 50.0\% mean success, exceeding $\pi_0$-Stateless by 43.6 percentage points and $\pi_0$-FrameStack by 35.2 points; its advantage over the stateless policy is positive across all five tasks. These comparisons support the value of maintaining a recurrent memory across policy calls rather than ignoring history or presenting a bounded history directly in context. \method\ also surpasses $\pi_0$-Vanilla Recurrent Memory by 33.2 points; as both policies maintain a recurrent state, this comparison indicates the merit of our proposed memory formulation beyond recurrence alone.

\begin{figure}[!htbp]
    \centering
    \begin{subfigure}[t]{0.55\linewidth}
        \centering
        \includegraphics[width=\linewidth]{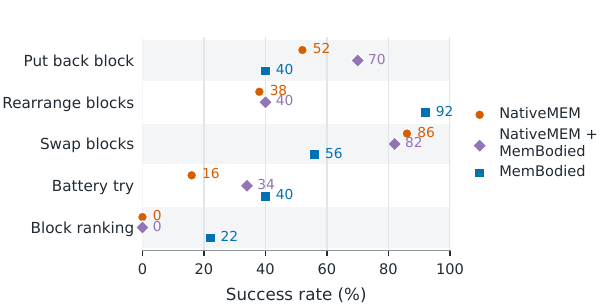}
        \caption{Success on each RMBench task.}
        \label{fig:nativemem_tasks}
    \end{subfigure}\hfill
    \begin{subfigure}[t]{0.44\linewidth}
        \centering
        \includegraphics[width=0.8\linewidth]{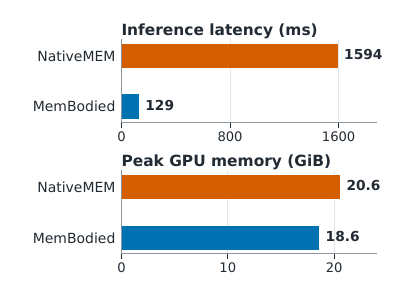}
        \caption{Inference Efficiency Comparison.}
        \label{fig:nativemem_efficiency}
    \end{subfigure}
    \caption{{NativeMEM and \method\ in the same evaluation setup.} We evaluate NativeMEM's video-history encoding alone and combined with \method, alongside standalone \method. Standalone \method\ exceeds NativeMEM by 11.6 percentage points on average.}
    \label{fig:nativemem_comparison}
    \vspace{-1.5em}
\end{figure}

\paragraph{Comparison with compressed history methods.}
Following \cref{fig:nativemem_comparison}, even with this more granular access to history, NativeMEM reaches 38.4\% mean success, compared with 50.0\% for standalone \method. Adding \method\ to NativeMEM improves its mean to 45.2\%, but remains 4.8\% below standalone \method. The gain from adding \method\ to NativeMEM could reflect more focused memory retrieval by \method. However, NativeMEM's history tokens could also draw attention away from the information retrieved by \method, potentially explaining why the combined model performs worse than standalone \method. Exploring how to combine explicit video history with associative memory more effectively is a direction for future work.

However, as seen in \cref{fig:nativemem_tasks}, no single method dominates at the task level. NativeMEM is stronger than standalone \method\ on Put Back Block and Swap Blocks, whereas \method\ is stronger on Rearrange Blocks, Battery Try, and Block Ranking. The combined model obtains the highest Put Back Block rate, but does not surpass the strongest standalone methods on the other four tasks. These results favour \method\ in the five-task mean success rate while showing that compressed history and associative memory have complementary, task-dependent strengths.

\begin{figure}[!htbp]
    \centering
    \begin{subfigure}[t]{0.49\linewidth}
        \centering
        \includegraphics[width=0.8\linewidth]{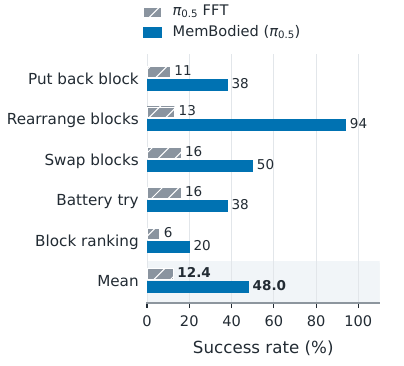}
        \caption{$\pi_{0.5}$ FFT baseline and \method.}
        \label{fig:pi05_success}
    \end{subfigure}\hfill
    \begin{subfigure}[t]{0.49\linewidth}
        \centering
        \includegraphics[width=0.8\linewidth]{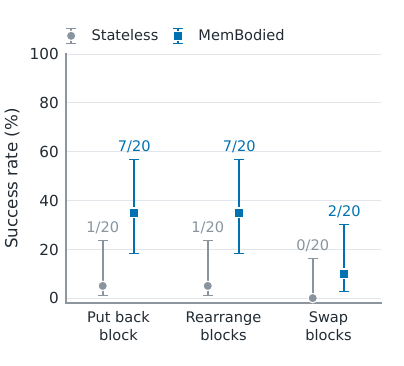}
        \caption{Real robot Evaluation. Whiskers are marginal 95\% Wilson binomial confidence intervals.}
        \label{fig:real_robot_eval}
    \end{subfigure}
    \caption{{Additional backbone and real-robot memory tasks.}}
    \label{fig:transfer}
\end{figure}

\paragraph{The gains transfer to $\pi_{0.5}$ backbone.} With $\pi_{0.5}$ backbone (\cref{fig:pi05_success}), \method\ reaches 48.0\% mean success as compared to 12.4\% for the $\pi_{0.5}$ baseline, improving on all five evaluated tasks. The largest gain is achieved on Rearrange Blocks, from 13.0\% to 94.0\%; Put Back Block improves from 11.0\% to 38.0\%. 

\paragraph{\method\ improves real-robot performance.} As shown in \cref{fig:real_robot_eval}, \method\ increases mean success across three real-robot tasks from 3.33\% to 26.67\%, with improvements on every task. 

\subsection{Inference efficiency}

As shown in \cref{fig:nativemem_efficiency}, \method{} achieves 91.9\% lower inference latency while using 9.5\% less peak GPU memory, as compared to NativeMEM. \method\ also introduces substantially fewer 40M parameters (only 1.26\% of $\pi_0$), roughly one-tenth of NativeMEM's 415M (12.81\% of $\pi_0$). During inference, NativeMEM encodes the incoming observations with its video-history encoder and adds the resulting tokens to the VLA context, introducing an additional encoding step and a growing context size. In contrast, \method\ maintains a fixed-size recurrent state that is read during action generation and updated at the start of a new policy call. Together, these results show that \method\ achieves higher mean success than NativeMEM with substantially lower inference latency, parameter overhead, and memory footprint.

\subsection{Performance on fully-observable manipulation}

\begin{wraptable}{r}{0.5\linewidth}
    \vspace{-1.5em}
    \centering
    \caption{\textbf{LIBERO success rates (\%).}} 
    \vspace{-0.3em}
    \label{tab:libero}
    \small
    \setlength{\tabcolsep}{5pt}
    \renewcommand{\arraystretch}{1.08}
    \resizebox{\linewidth}{!}{%
    \begin{tabular}{@{}lrrrrr@{}}
        \toprule
        Method & Spatial & Object & Goal & Long & Mean \\
        \midrule
        Diffusion Policy~\citep{chi2025diffusion} & 78.3 & 92.5 & 68.3 & 50.5 & 72.4 \\
        Octo~\citep{team2024octo} & 78.9 & 85.7 & 84.6 & 51.1 & 75.1 \\
        OpenVLA~\citep{pmlr-v270-kim25c} & 84.7 & 88.4 & 79.2 & 53.7 & 75.9 \\
        SpatialVLA~\citep{qu2025spatialvla} & 88.2 & 89.9 & 78.6 & 55.5 & 78.1 \\
        UniACT~\citep{zheng2025universal} & 77.0 & 87.0 & 77.0 & 70.0 & 76.8 \\
        $\pi_0$~\citep{black2024pi_0} & \textbf{96.8} & \textbf{98.8} & \textbf{95.8} & 85.2 & 94.2 \\
        $\pi_0$-FAST~\citep{pertsch2025fast} & 96.4 & 96.8 & 88.6 & 60.2 & 85.5 \\
        \midrule
        \method & 96.2 & 97.8 & \textbf{95.8} & \textbf{90.6} & \textbf{95.1} \\
        \bottomrule
    \end{tabular}
    }
    \vspace{-10pt}
\end{wraptable}

\paragraph{The largest improvement is skewed toward long-horizon LIBERO tasks.} LIBERO provides a complementary setting for RQ3 by testing \method\ on fully observable tasks where episodic recall is not required to complete the task. \Cref{tab:libero} shows that \method\ reaches 90.6\% success on LIBERO-Long, exceeding $\pi_0$'s score of 85.2\% by 5.4 percentage points. In contrast, Goal is unchanged, while Spatial and Object decrease by 0.6 and 1.0 points, respectively. The four-suite mean remains comparable at 95.1\% against $\pi_0$'s 94.2\%. The skewness of the improvement toward LIBERO-Long supports the view that \method\ is particularly useful for improving task performance on longer fully-observable manipulation, without degrading aggregate performance.

\section{Analyses}
\label{sec:analysis}

\begin{figure}[!htbp]
    \centering
    \begin{subfigure}[t]{0.54\linewidth}
        \centering
        \includegraphics[width=0.8\linewidth]{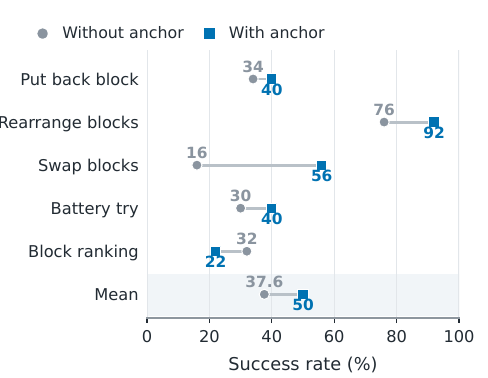}
        \caption{Fixed episode anchor.}
        \label{fig:anchor_impact}
    \end{subfigure}\hfill
    \begin{subfigure}[t]{0.45\linewidth}
        \centering
        \includegraphics[width=0.8\linewidth]{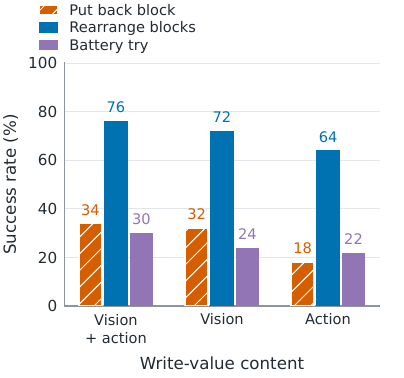}
        \caption{Write-value composition.}
        \label{fig:memory_value_ablation}
    \end{subfigure}
    \caption{{Memory design ablations on RMBench.}
    }
    \label{fig:memory_ablations}
\end{figure}

\paragraph{Memory architecture ablation.} \Cref{fig:anchor_impact,tab:rmbench_results} compare the associative-state variants and the contribution of the episode anchor pathway. Among the anchor-free variants, memory-token injection reaches 37.6\% mean success, compared with 23.2\% for hierarchical memory and 20.8\% for attention steering. This indicates that feeding the retrieved vector as contextual content is more effective, in our setting, than the two alternative memory designs. The combined two-pathway memory reaches 50.0\%, compared with 37.6\% when the anchor pathway is removed. Relative to the anchor-free variant, the \method improves on four out of five tasks, with the largest gain on Swap Blocks, from 16.0\% to 56.0\%. However, the anchor reduces performance on Block Ranking from 32.0\% to 22.0\%; this task depends more on tracking evolving task progress than recalling the initial scene, making the fixed first-frame reference potentially less useful. Overall, both the memory-token design and episode anchor contribute to the complete memory formulation, although the anchor's effect is task-dependent.

\paragraph{Vision and action play complementary roles through memory.} Across the three evaluation tasks, \cref{fig:memory_value_ablation} shows that combining the subsequent visual observation with the action summary achieves 46.7\% mean success, compared with 42.7\% for vision-only and 34.7\% for action-only memory. On Put Back Block and Rearrange Blocks, visual information accounts for most of the measured performance, with the action summary adding 2 and 4 percentage points. On Battery Try, the combined value reaches 30.0\%, compared with 24.0\% for vision alone and 22.0\% for action alone; removing either modality therefore reduces success by 6 or 8 points. The combined representation is the strongest on every task, indicating the complementarity of these two channels.

\paragraph{Qualitative analysis on the impact of memory.} Manual annotation of 50 seed-matched rollout pairs per task shows that \method\ and $\pi_0$-Stateless behave similarly when the current observation is sufficient, but diverge at decisions requiring past information. For example, on Put Back Block, \method\ reduces wrong-pad returns from 37/50 to 13/50, although both policies complete the preceding block movement and button press in every rollout (see more details in Appendix~\ref{sec:qualitative_analysis}). Across tasks, \method\ more often selects the episode-specific target, tracks task progress, and uses earlier outcomes to guide subsequent attempts. Its remaining failures largely involve motor execution or longer-term tracking. These observations localise \method's gains to decisions that require information no longer visible in the current observation.

\FloatBarrier

\section{Conclusion}

In this work, we introduced \method, a fixed-capacity episodic memory that combines an associative memory state with an initial-scene anchor without expanding the policy context. Across five memory-dependent RMBench tasks, it achieved 50.0\% mean success, outperforming stateless, frame-history, vanilla recurrent, and compressed-history baselines. The improvements also extended to the $\pi_{0.5}$ backbone and three real-robot tasks. Importantly, the gains in memory-dependent manipulation do not come at the expense of inference efficiency. Compared with NativeMEM, \method\ pairs higher mean success with substantially lower latency and memory footprint. \method\ also preserves general manipulation ability, maintaining comparable mean success on LIBERO while improving on the long-horizon suite. Together, these results support \method\ as a practical alternative to expanding observation histories in VLA control.

\subsection*{AI use statement}

In this work, we used generative AI tools to provide feedback on research methodology and experiments, assist with method implementation, create or modify scientific figures, summarise existing literature, and draft or edit portions of the manuscript for clarity. We did not use these tools to develop theoretical models, formulate or prove mathematical claims, propose or refine hypotheses, clean or reformat datasets, generate synthetic data, or translate the manuscript. We manually checked LLM-generated code for correctness, verified literature summaries against the cited sources, and reviewed all AI-generated text and figures. We take responsibility for the final content of this work, including text, claims or artifacts produced with the aid of generative AI.

\bibliography{iclr2027_conference}
\bibliographystyle{iclr2027_conference}

\appendix

\section{Extended Related Work}
\label{app:extended_related_work}

This section expands the review in Section~\ref{sec:related_works}, providing additional context on the broader literature surrounding generalist VLA policies, memory-augmented manipulation, associative memory, and memory-focused evaluation.

\subsection{Generalist VLA policies}

Vision-language-action models transfer semantic representations from pretrained vision-language models into robot control. RT-1 consumes a bounded image history and compresses visual tokens before causal processing \cite{brohan2022rt}. RT-2 and OpenVLA represent robot actions as tokens within vision-language backbones \cite{brohan2023rt,pmlr-v270-kim25c}. Octo instead trains a diffusion-based generalist policy from heterogeneous robot trajectories \cite{team2024octo}. $\pi_0$ combines a pretrained PaliGemma vision-language model with a flow-matching action expert for continuous action chunk prediction \cite{black2024pi_0}. OpenVLA-OFT shows that continuous action representations, parallel decoding, and action chunking can improve control speed and success, while $\pi_{0.5}$ combines low-level control with semantic prediction and heterogeneous co-training for open-world tasks \cite{kim2025fine, intelligence2025pi_}.

Across these architectures, temporal context is limited to a bounded observation history or an action chunk. Such local context can preserve recently visible evidence, but it does not provide a persistent episodic state for information that must be retained beyond the input window or selectively carried across an episode.

\subsection{Memory representations for manipulation}

Recent memory-aware policies can be distinguished by the representation they preserve. \textbf{Compressed-history methods} retain selected observations as tokens. HAMLET~\cite{koo2026hamlet} encodes observations into moment tokens and aggregates them through a lightweight history module. NativeMEM~\cite{wang2026nativemem} reuses the VLA vision encoder to compress each historical camera frame into a native memory token. These methods preserve explicit temporal evidence more efficiently than raw frame histories, although their context can still grow with the number of retained observations.

\textbf{Fixed-size recurrent methods} compress history into a bounded latent state. $\mu$VLA carries learnable memory tokens across OpenVLA-OFT control steps and trains them with truncated backpropagation through time \cite{cherepanov2026mu}. ReMem-VLA propagates separate frame and chunk level recurrent queries \cite{li2026remem}, while AVA-VLA uses recurrent state to modulate current visual features \cite{xiao2026ava}.

\textbf{Memory-bank and retrieval methods} provide more structured storage. SAM2Act~\cite{fang2025sam2act} integrates a visual memory architecture for manipulation, and MemoryVLA combines working-memory tokens with a bank that preserves visual detail and higher-level semantics \cite{shi2026memoryvla}. MAP-VLA~\cite{li2025map} retrieves task-stage prompts learned from demonstrations. Notes-to-Self~\cite{haresh2026self} externalises object locations, plans, and subgoal progress in a language scratchpad. These systems make memory content more explicit, but they introduce a bank, retrieval rule, or language-generation channel in addition to the recurrent policy state.

\method\ occupies a different position in this design space. It neither retains a growing sequence of observations nor retrieves discrete memory entries. Instead, it represents episode history through a set of layer-wise associative matrices and an initial-scene anchor, both with storage independent of episode length, and conditions the action expert on their readouts.

\subsection{Fast-weight associative memory}

Fast-weight models separate slowly optimised parameters from rapidly updated episode state. \citet{6796337} introduced learned control of fast-weight memories. Linear attention was later interpreted as an associative fast-weight programmer \cite{schlag2021linear}. DeltaNet replaced purely additive key-value writes with a delta rule that corrects the association already recalled for a key, and Gated DeltaNet added data-dependent retention \cite{yang2025gated}. Most directly related, $\delta$-mem augments a language model with a fixed-size online matrix whose readout produces low-rank attention corrections \cite{lei2026delta}.

\method\ transfers the associative-state principle to continuous robot control. Rather than perturbing attention queries or outputs, it injects a projected memory vector into a dedicated suffix token before attention. We use this interface comparison to analyse how an associative state is most effectively exposed to the action network.

\subsection{Memory-focused evaluation}

Memory-specific manipulation benchmarks have recently proliferated. MemoryBench evaluates spatial memory and action recall, requiring policies to recover scene information or actions from earlier in an episode \cite{fang2025sam2act}. MIKASA-Robo broadens this scope through a diverse set of partially observable tabletop tasks designed to test memory-intensive control \cite{cherepanov2026memory}. RMBench comprises nine dual-arm manipulation tasks organised by memory complexity, ranging from retaining information from a single earlier stage to accumulating evidence across multiple observations or attempts \cite{chen2026rmbench}. We use RMBench as our primary memory evaluation benchmark because its controlled $M(1)$ and $M(n)$ categories distinguish tasks that require retaining a single earlier event from those that require accumulating information across several events within a common bimanual manipulation setting.  In contrast, LIBERO is not designed specifically around episodic recall; its spatial, object, goal, and long-horizon suites provide a complementary test of whether a memory-augmented policy remains effective on general manipulation \cite{liu2023libero}.

\section{Additional Experimental Details}
\label{app:implementation_details}

This section describes the evaluated tasks, training configurations, baseline implementations, and inference-efficiency profiling. Table~\ref{tab:common_training_config} summarises the shared training configurations and hyperparameters.

\begin{table}[!htbp]
    \centering
    \caption{\textbf{Training configuration and Hyperparameters.}}
    \label{tab:common_training_config}
    \small
    \setlength{\tabcolsep}{5pt}
    \renewcommand{\arraystretch}{1.08}
    \begin{tabular}{@{}p{0.29\linewidth}p{0.65\linewidth}@{}}
        \toprule
        Setting & Configuration \\
        \midrule
        Visual input & Three camera streams, resized with padding to $224\times224$ \\
        Action generation & 50-step action horizon; 10 flow-matching denoising steps \\
        Numerical settings & bfloat16; seed 42; EMA 0.99; global gradient clipping at 1 \\
        Optimiser & AdamW, $(\beta_1,\beta_2)=(0.9,0.95)$, $\epsilon=10^{-8}$, weight decay $10^{-10}$ \\
        Learning-rate schedule & Cosine decay with 1,000 warm-up steps, peak rate $2.5\times10^{-5}$, and terminal rate $2.5\times10^{-6}$ \\
        \bottomrule
    \end{tabular}
\end{table}

\subsection{Evaluated RMBench tasks}

Our evaluation uses five RMBench tasks that cover both memory-complexity categories~\cite{chen2026rmbench}: three $M(1)$ tasks and two $M(n)$ tasks.

\paragraph{$M(1)$ tasks.} These tasks require information from a single prior event in the observations. In \textbf{Put Back Block}, the robot moves a block from one of four pads to the centre, presses a button, and returns it to its original pad, requiring it to remember which pad was initially occupied. In \textbf{Rearrange Blocks}, the robot moves the block between two pads onto the empty pad, presses a button, and then moves the block that originally occupied a pad into the middle, requiring it to retain the initial layout. In \textbf{Swap Blocks}, the robot exchanges two blocks placed on two of three trays using the empty tray as temporary space, requiring it to remember the initial block-tray assignment and block position throughout the transfers.

\paragraph{$M(n)$ tasks.} These tasks require accumulating information across multiple prior events which are specifically attempts in this case. In \textbf{Battery Try}, the robot inserts two batteries in different orientations until the dashboard indicates the correct combination, requiring it to remember which combinations have already been tested. In \textbf{Block Ranking}, the robot rearranges three coloured blocks and presses a button to test each ordering until the correct one is found, requiring it to remember which orderings have already failed.

\subsection{RMBench configuration}

The RMBench models adapted the PaliGemma backbone using LoRA rank 16 and alpha 16, and the action expert using rank 32 and alpha 32. Models were trained for 10,000 steps with batch size 8. Gradients propagated through the complete sampled sequence. Consecutive sequence elements were separated by 50 environment steps, matching one action horizon. The sequence length was 8 for Put Back Block, Rearrange Blocks, and Swap Blocks; 14 for Battery Try; and 18 for Block Ranking, determined based on average episode lengths in the task. he main \method\ configuration used associative rank $r=128$ and $\alpha_{\mathrm{mem}}=256$, giving an explicit retrieval multiplier of $\alpha_{\mathrm{mem}}/r=2$.

\subsection{LIBERO Configuration}

We trained the LIBERO experiment differently from the RMBench models using full fine-tuning to match the published $\pi_{0}$ baseline setup. Training used a batch size of 32, 30,000 steps, and a sequence length of six.  Evaluation used five-step replanning, following the official $\pi_{0}$ LIBERO setup. To match this replanning cadence while retaining the 50-step memory-update interval used during training, we maintained ten recurrent memory slots in a round-robin schedule. Each five-step policy call reads one slot, and that slot was revisited after 50 environment steps. At the revisit, the delayed write used the new observation together with a summary of the complete 50-step action chunk predicted on the slot's previous visit. This summary therefore includes predicted actions beyond the five steps executed before the next replan.

\subsection{Baseline implementations}

Table~\ref{tab:baseline_implementation} records how each evaluated baseline was implemented. Diffusion Policy, ACT, X-VLA, and the $\pi_{0.5}$ baseline in Table~\ref{tab:rmbench_results} are published RMBench results~\citep{chen2026rmbench}.

\begin{table}[!htbp]
    \centering
    \caption{\textbf{Implementation details for locally evaluated baselines.} ``Local adaptation'' indicates that the method was implemented and trained within our $\pi_0$ evaluation setup.}
    \label{tab:baseline_implementation}
    \small
    \setlength{\tabcolsep}{4pt}
    \renewcommand{\arraystretch}{1.08}
    \begin{tabular}{@{}p{0.18\linewidth}p{0.16\linewidth}p{0.6\linewidth}@{}}
        \toprule
        Method & Type & Implementation \\
        \midrule
        $\pi_0$-Stateless & Local baseline & Current observation only; batch size 32; 10,000 steps. \\
        $\pi_0$-FrameStack & Local baseline & Four observations separated by 50 environment steps and concatenated in the policy context; samples with insufficient history were padded to four frames. \\
        $\pi_0$-$\mu\text{-VLA}$ & Local adaptation & 64 recurrent memory tokens; sequence length 8; stride 1; gradients truncated every two recurrent steps. \\
        $\pi_0$-Vanilla\newline Recurrent Memory & Local adaptation & Rank 128 and scale 256; query/output attention corrections; recurrent hidden-state writes. \\
        NativeMEM & Local adaptation & Memory-tokenizer training for 50,000 steps, offline token caching, then policy training for 20,000 steps with queue stride 1. \\
        NativeMEM\newline + \method &Combination & NativeMEM pipeline above with \method{} sequence length 8 and sequence stride 50. \\
        $\pi_0$-Hint & Inference-only reference & Simulator-derived textual task-progress hints appended at inference, without retraining. \\
        \bottomrule
    \end{tabular}
\end{table}

The $\pi_0$-Hint annotations exposed only task progress relevant to the evaluated memory demands. For Put Back Block, they identified the block's original mat and whether it had been moved, the button pressed, and the block returned. For Rearrange Blocks, they recorded whether the first block had reached the empty mat, the button had been pressed, and the second block had been moved between the mats. For Swap Blocks, they described completion of the three block-transfer stages and the button press. Battery Try listed previously attempted polarity combinations and whether the dashboard was on, while Block Ranking listed colour orders attempted before earlier button presses. Because the stateless policy was not trained to consume these annotations, $\pi_0$-Hint is a privileged-information reference rather than an upper bound.

The internal variants changed how the associative state affected action generation. \method{}-AS projected each retrieval into additive corrections to the attention query and output representations. \method{}-H augmented the gated-delta state with a slower LSTM-like recurrent cell controlled by a learned forget gate. The latter therefore changed the state dynamics as well as its use by the action network, so these results compare complete memory designs rather than isolated readout substitutions.

Unlike the other locally trained RMBench baselines, the NativeMEM baseline followed a separate staged training procedure. A memory tokenizer was first trained on the RMBench mixture for 50,000 steps, after which its per-frame, per-view tokens were cached offline for each target task. The tokenizer was then frozen while the queue-conditioned policy was trained for 20,000 steps. The standalone model used a queue stride of one; the combined NativeMEM + \method{} model retained this queue while training the associative state with sequence length 8 and stride 50.

\subsection{memory variants}

The reported attention-steering and hierarchical variants use the same associative memory configuration as \method.

\paragraph{Attention steering (\method{}-AS).} At each action-expert layer, the state token queries the associative matrix. Separate learned projections turn the readout into additive corrections to the action tokens' attention queries and projected attention outputs. The readout is scaled by $\alpha_{\mathrm{mem}}/r$, and each correction is scaled by the root-mean-square magnitude of the representation it modifies.

\paragraph{Hierarchical memory (\method{}-H).} This variant retains the attention-steering readout but adds a second, slower matrix $C_t^{(l)}$ to each layer's recurrent state. The gated-delta write produces an updated associative matrix for each value slot. A sigmoid forget gate derived from the post-layer state token blends their average with the previous slow matrix $C_{t-1}^{(l)}$ to obtain $C_t^{(l)}$. The final associative state is a weighted average of each gated-delta matrix and $C_t^{(l)}$. Thus, \method{}-H changes the state update as well as the readout relative to the main memory-token model; the variants are complete designs, not isolated readout substitutions.

\subsection{$\pi_{0.5}$ adaptation}

Unlike $\pi_0$, $\pi_{0.5}$ represents robot state in the language prefix and has no dedicated state token in the action-expert suffix. We therefore use the memory token's hidden representation for both associative-memory operations. At each action-expert layer, the token's pre-attention representation is projected to form the read query, and the gated retrieval is then added to that token before attention. Its post-layer representation is projected to form the write key for the subsequent memory update, whose value combines the action summary with the following visual observation. The token is recreated on each call, whereas the associative matrices persist across calls within an episode. The initial-scene anchor is retained by fusing its readout into the action tokens.

\subsection{Inference-efficiency profiling}

We profiled \method\ and NativeMEM in bfloat16 on NVIDIA A100 GPUs with 80\,GB of VRAM, using one GPU per evaluation run. We measured model-side computation only. For \method, timing covered policy calls from invocation until the action array was returned. This includes policy-side input preprocessing, current-observation vision encoding, associative-memory reads and updates, and generation of a 50-step action chunk using 10 flow-matching denoising steps. For NativeMEM, per-cycle latency was the sum of the model-side time for all video history encoding calls since the preceding policy query and the subsequent policy inference call. This includes per-frame video-history encoding, updates to the expanding history queue and policy context, and action generation.

We evaluated Put Back Block, Rearrange Blocks, Swap Blocks, and Battery Try for 50 episodes per task and policy, yielding 200 episodes per policy. The first three policy queries of each task-policy run were treated as warm-up and excluded. The reported steady-state latency is the arithmetic mean over all retained policy cycles pooled across the four tasks, comprising 2,548 NativeMEM cycles and 2,563 \method{} cycles. Peak memory is the maximum JAX allocator \texttt{peak\_bytes\_in\_use} observed across all calls and tasks. NativeMEM required 1,593.7\,ms per policy cycle and 20.57\,GiB of peak GPU memory, whereas \method\ required 129.2\,ms and 18.61\,GiB, corresponding to 91.9\% lower latency and 9.5\% lower peak memory.

\section{Real-World Data and Robot Evaluation}
\label{app:real_world}

\subsection{Robot Details}

\begin{figure}[!htbp]
    \centering
    \begin{subfigure}[b]{0.22\linewidth}
        \centering
        \includegraphics[height=3.1cm]{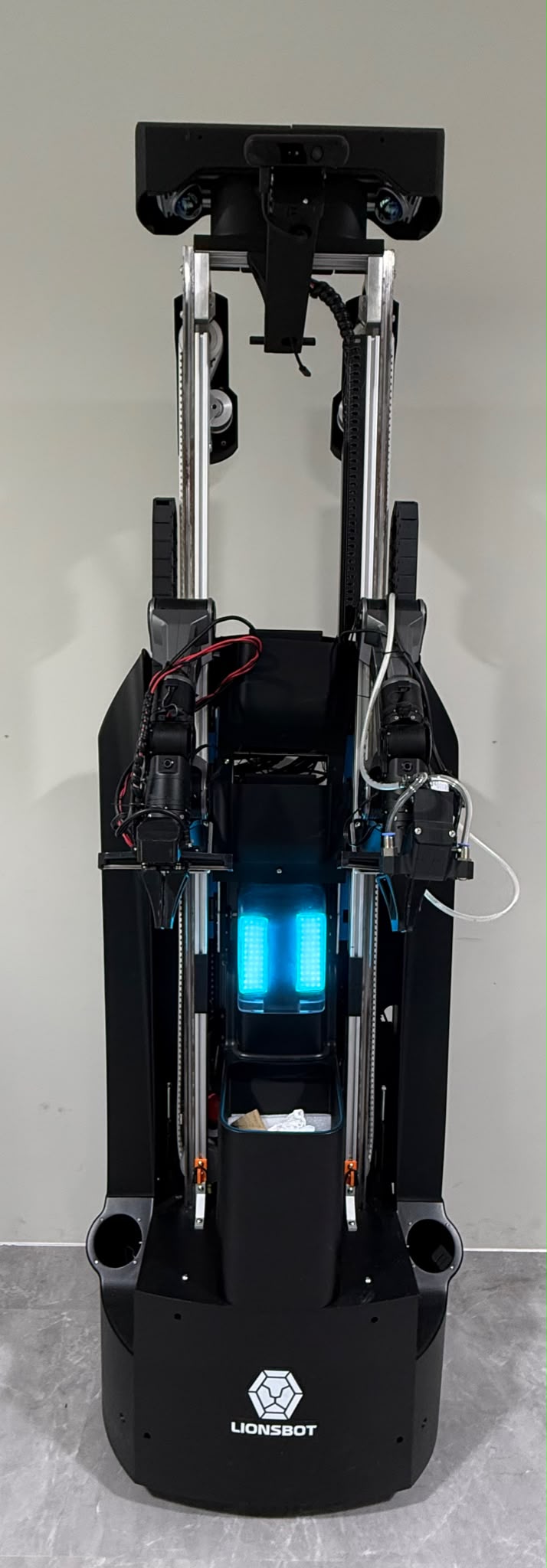}
        \caption{Robot}
    \end{subfigure}\hfill
    \begin{subfigure}[b]{0.237\linewidth}
        \centering
        \includegraphics[width=\linewidth]{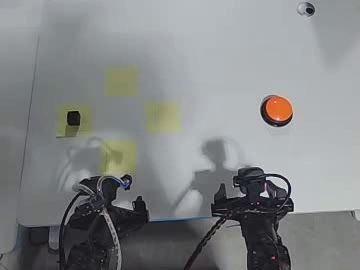}
        \caption{Put Back Block.}
    \end{subfigure}\hfill
    \begin{subfigure}[b]{0.245\linewidth}
        \centering
        \includegraphics[width=\linewidth]{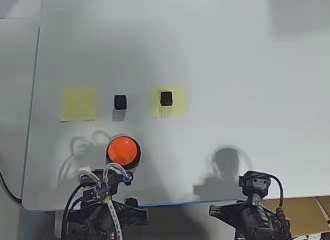}
        \caption{Rearrange Blocks.}
    \end{subfigure}\hfill
    \begin{subfigure}[b]{0.245\linewidth}
        \centering
        \includegraphics[width=\linewidth]{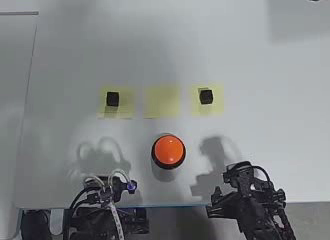}
        \caption{Swap Blocks.}
    \end{subfigure}
    \caption{\textbf{Real-robot platform and representative task configurations.} The task panels show top-camera observations approximately one second after the start of a cleaned demonstration. They illustrate the initial workspace configurations and are not evaluation rollouts.}
    \label{fig:real_robot_setup}
\end{figure}

The robot comprised two AgileX PiPER arms, an Orbbec Gemini 336L top camera, and two Intel RealSense D405 wrist cameras (Figure~\ref{fig:real_robot_setup}). The robot state and action vectors each contained 14 values: six joint values and one gripper value for each arm.

\subsection{Dataset collection and preprocessing}

We manually collected 50 demonstrations for each of the three tasks. Each demonstration contained synchronised top, left-wrist, and right-wrist views recorded at 30~Hz, together with robot states and actions. Table~\ref{tab:real_robot_data} summarises the data retained after cleaning. The cleaned top-camera videos have resolution $330\times240$, while both wrist-camera streams have resolution $424\times240$.

\begin{table}[!htbp]
    \centering
    \caption{\textbf{Real-world demonstration data after cleaning.} Mean duration is computed from the retained 30~Hz trajectories.}
    \label{tab:real_robot_data}
    \small
    \setlength{\tabcolsep}{6pt}
    \renewcommand{\arraystretch}{1.08}
    \begin{tabular}{@{}lrrrr@{}}
        \toprule
        Task & Collected & Retained & Frames & Mean duration (s) \\
        \midrule
        Put Back Block & 50 & 41 & 54,228 & 44.1 \\
        Rearrange Blocks & 50 & 50 & 48,914 & 32.6 \\
        Swap Blocks & 50 & 49 & 86,818 & 59.1 \\
        \bottomrule
    \end{tabular}
\end{table}

Demonstrations affected by recording faults were removed before training. The retained trajectories were temporally trimmed to the demonstrated interaction, and the recorded joint states were denoised. The top-camera stream was cropped to the task workspace during preprocessing. Task-specific real-robot policies were trained on the corresponding cleaned datasets using a training recipe similar to the RMBench experiments.

\subsection{Evaluation protocol}

We evaluated each policy for 20 physical rollouts on each task. A rollout was counted as successful only when the complete instructed task was achieved, and partial completion received no credit. The recurrent associative state and initial-scene anchor were reset at the start of every rollout. The resulting success rates are reported in Figure~\ref{fig:real_robot_eval}.

\section{Additional Analysis}

\subsection{Additional baseline Clarification}

Table~\ref{tab:rmbench_results} shows that our $\pi_0$-$\mu$VLA adaptation achieved 0.0\% success on all five RMBench tasks under the shared training budget. This result shows only that the adaptation did not learn an effective recurrent policy under our training recipe; it should not be interpreted as a general comparison with $\mu$VLA under its original setting. One possible explanation for the poor performance in our adaptation is that learning recurrence through carried tokens may require longer training. The inference-time $\pi_0$-Hint policy reached 2.0\% mean success. Because the policy was not trained to consume its
simulator-derived textual hints, this condition is a privileged-information reference rather than a trained oracle or an upper bound.

\subsection{Memory Rank}

\begin{figure}[!htbp]
    \centering
    \includegraphics[width=0.4\linewidth]{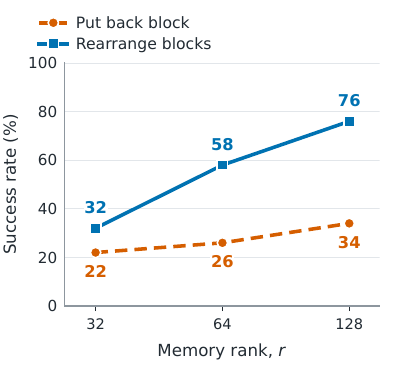}
    % \caption{Associative memory rank.}
     \caption{Associative memory rank comparison in the anchor-free memory-token variant.}
    \label{fig:memory_rank_ablation}
\end{figure}

In the anchor-free memory-token variant, success increases monotonically as the rank grows from 32 to 64 to 128 on both Put Back Block and Rearrange Blocks (Figure~\ref{fig:memory_rank_ablation}). Their two-task mean rises from 27.0\% to 42.0\% and 55.0\%, respectively. The policy's sensitivity to rank is consistent with associative-state capacity influencing action selection: a larger state can preserve a richer set of episode-dependent associations for later retrieval.

\subsection{Initial-Frame Anchor Representation}

We also tested a simpler way to provide the initial scene: adding the episode's first frame to the policy prefix alongside the current input, rather than using the fixed episode-anchor pathway. On Put Back Block task, this first-frame-prefix variant achieved 22.0\% success, compared with 34.0\% for the memory-token model without an anchor and 40.0\% with the proposed anchor. This could be because adding the full first frame introduces distracting visual tokens into the policy context, whereas the anchor selectively retrieves a compact reference to the initial scene.

\subsection{Cross-Episode Memory Carryover}

\begin{table}[!htbp]
    \centering
    \caption{RMBench success rates (\%) with associative memory reset or carried across episodes. The initial-frame anchor is reset at the start of every episode in both conditions.}
    \label{tab:cross_episode_memory}
    \small
    \begin{tabular}{@{}lrr@{}}
        \toprule
        Task & Memory reset & Memory carried over \\
        \midrule
        Put Back Block & 40.0 & 18.0 \\
        Rearrange Blocks & 92.0 & 78.0 \\
        Swap Blocks & 56.0 & 36.0 \\
        Battery Try & 40.0 & 24.0 \\
        Block Ranking & 22.0 & 28.0 \\
        \midrule
        Mean & 50.0 & 36.8 \\
        \bottomrule
    \end{tabular}
\end{table}

As an exploratory test, we carry the associative state across rollout boundaries, although \method\ was trained with the state reset at the start of each episode. The initial-frame anchor is still reset for each new rollout. Table~\ref{tab:cross_episode_memory} shows that this reduces mean success from 50.0\% to 36.8\%, with lower success on four of the five tasks. The largest declines occur on Put Back Block and Swap Blocks (22 and 20 percentage points), where the relevant location or progress must be established within the current episode. Block Ranking is the exception, improving by 6 points. Its trial-and-error structure may respond differently to carried state, although this increase does not establish that information from earlier episodes was useful. Overall, the results do not support a general benefit from carrying over memory under this evaluation, as the declines are consistent with information from earlier episodes interfering with episode-specific decisions. 

\section{Qualitative Analysis}
\label{sec:qualitative_analysis}

The mean success rates show that \method\ improves memory-dependent control, but do not identify whether the gains arise at the decisions that require history. We therefore
manually annotate all 50 paired rollouts of $\pi_0$-Stateless and \method, using identical seeds, and identify the stages at which their behaviours diverge. The analysis separates RMBench tasks that require recalling one earlier observation, denoted $M(1)$, from tasks that require accumulating the outcomes of several attempts, denoted $M(n)$. Across both groups, the policies behave similarly when the current observation is sufficient, but diverge once the correct action depends on information that is no longer visible.

\subsection{Recalling an earlier observation}

\begin{figure}[t]
    \centering
    \includegraphics[width=\linewidth]{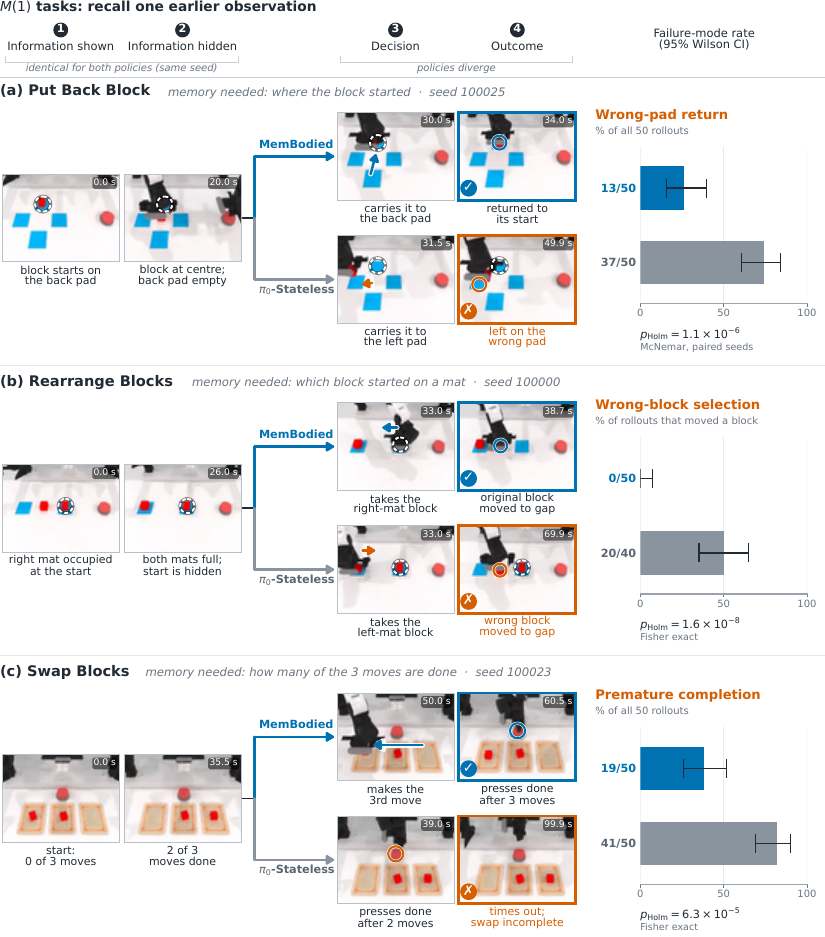}
    \caption{\textbf{Failure modes on the three $M(1)$ tasks.} Representative paired rollouts show the same seed as task-relevant information becomes hidden and the policies reach the ambiguous decision. The blue trajectory corresponds to \method\ and the grey trajectory to $\pi_0$-Stateless. Bars summarise the annotated failure mode across rollouts, with marginal 95\% Wilson confidence intervals. Put Back Block and Swap Blocks use all 50 rollouts per policy; Rearrange Blocks is conditioned on rollouts in which a block is moved (50 for \method, 40 for $\pi_0$-Stateless). The $p$-values shown in the figure are Holm-corrected.}
    \label{fig:failure_modes_m1}
\end{figure}

\subsubsection{Put Back Block}
Put Back Block isolates spatial recall: both policies move the block to the centre and press the button in every rollout, but they differ when deciding where to return it. The stateless policy returns to the wrong pad in 37/50 rollouts. Its 13 correct choices are close to the 13.8 expected from its own pad preference, indicating that its return location does not track the episode's initial pad. \method\ reduces wrong-pad returns to 13/50 and corrects 24 stateless errors without introducing the reverse error on any paired seed. Its 37 correct choices also exceed the 14.5 expected by chance. Both comparisons are significant at $p{<}10^{-4}$. The benefit of \method\ is therefore concentrated on the history-dependent choice of return location, rather than the preceding pick-and-place sequence.

Aside from the recall errors, the remaining failures expose two execution constraints. Of the 17 correct returns towards the front pad, 14 reach the pad but stall instead of releasing the block. The stateless policy exhibits the same tendency in 8/9 relevant rollouts, suggesting a shared placement difficulty rather than a memory-specific failure. In addition, six of \method's eight failures classified as residual recall errors occur when the block starts on the right: the policy repeats its dominant left-arm motion towards the front pad instead of switching arms. These two patterns account for 20 of the 30 \method\ failures. Thus, memory largely resolves the target location recall problem, but not the motor strategy needed to reach every target reliably.

\subsubsection{Rearrange Blocks}
Rearrange Blocks tests recall through object selection rather than target selection. After the first placement and button press, both mats contain identical blocks, so only the initial layout identifies which block should be returned to the gap. Both policies complete the shared early stages in every rollout, but their subsequent choices differ. \method\ selects the correct block specified by the initial layout in all 50 rollouts. The stateless policy instead exhibits a strong left-block preference: it chooses the left block in 37/40 rollouts in which it moves a block, selects the correct block only 20 times, and moves no block in the remaining ten rollouts. This shows that memory changes the episode-specific decision made after the two scenes become visually indistinguishable, rather than merely improving the manipulation stages.

Object selection does not explain the entire performance gap. Of the 20 stateless rollouts that select the correct block and bring it to the target pad, only one succeeds, whereas \method\ succeeds in 46/50 rollouts overall. The stateless policy also does not press the button fully, leading to task failures. These execution errors occur after the history-dependent object selection and therefore compound, rather than explain, the recall gap.

\subsubsection{Swap Blocks}

Swap Blocks differs from the previous tasks because the hidden variable is progress rather than an initial location. Exchanging two identical blocks requires three moves through an empty pad, after which the visible occupancy matches the start of the episode. Intermediate occupancy patterns can also be valid starting layouts. Consequently, the current image does not reveal whether the policy should move another block or press the completion button.

Both policies can perform the required manipulation: \method\ makes at least one move in 48/50 rollouts and the stateless policy in 41/50. The main difference is in deciding when the swap is complete. The stateless policy presses the completion button prematurely in 41/50 rollouts, compared with 19/50 for \method, and completes all three moves in only eight rollouts, compared to 31 for \method. Among rollouts in which the button is pressed, \method\ waits until the swap is complete in 28/47 cases, whereas the stateless policy does so in only 6/47. All three progress-related comparisons remain significant ($p{<}10^{-4}$). This pattern is consistent with memory tracking progress through the move sequence. However, the tracking is not perfect: 12 of the 22 residual \method\ failures press after two moves, making the final transition the dominant remaining error.

\subsection{Accumulating outcomes across attempts}

\begin{figure}[t]
    \centering
    \includegraphics[width=\linewidth]{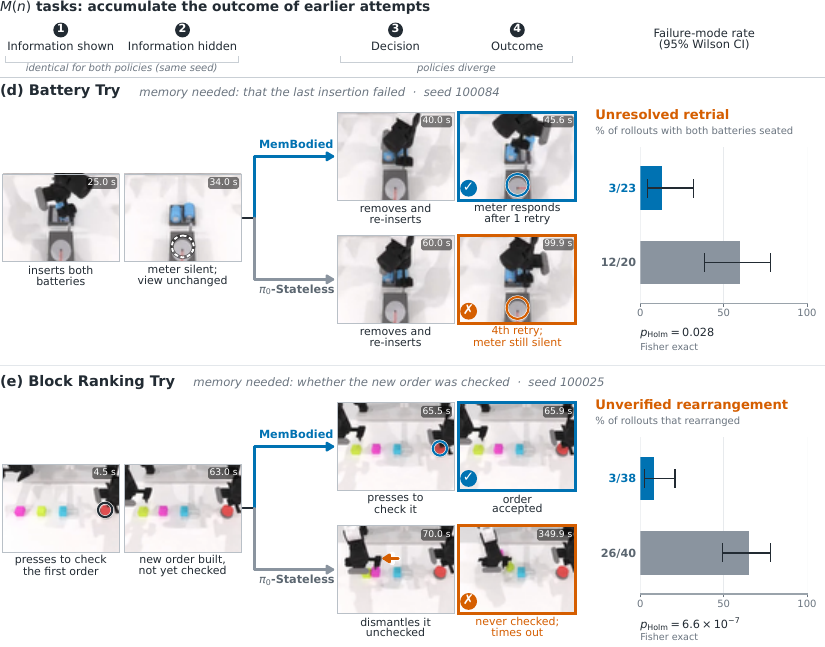}
    \caption{\textbf{Failure modes on the two $M(n)$ tasks.}
    Representative paired rollouts show decisions that depend on the outcome of an earlier attempt. The blue trajectory corresponds to \method\ and the grey trajectory to $\pi_0$-Stateless. Bars report the conditional failure rate with marginal 95\% Wilson confidence intervals: unresolved retries among rollouts that seat both batteries (23 for \method, 20 for $\pi_0$-Stateless), and unverified arrangements among rollouts that complete a rearrangement (38 and 40, respectively). The $p$-values shown in the figure are Holm-corrected.}
    \label{fig:failure_modes_mn}
\end{figure}

\subsubsection{Battery Try}
Battery Try requires the policy to change its next action after transient failure feedback. The two policies set both batteries at similar rates (23/50 for \method\ and 20/50 for $\pi_0$-Stateless), indicating comparable
grasp-and-insert behaviour before memory is needed. They diverge after an unsuccessful orientation test, once the meter has returned to rest and the current image no longer reveals what was tried. The stateless policy makes a median of four further attempts and completes the task in 8/20 rollouts; \method\ usually succeeds within one further attempt and resolves it in 20/23. This difference remains significant after Holm correction ($p{=}0.028$), localising the benefit of \method\ to the point where an earlier outcome must alter the next action. The dominant remaining limitation is grasp stability: 25 of the 30 \method\ failures involve dropping a battery during pickup or insertion, a common failure mode in both policies.

\subsubsection{Block Ranking Try}
Block Ranking Try extends the same dependence on earlier outcomes to a longer search. Both policies complete an initial block arrangement at similar rates (38/50 for \method\ and 40/50 for $\pi_0$-Stateless), but differ in whether they verify it. \method\ presses the test button after 35/38 completed arrangements, compared with 14/40 for the stateless policy. On four paired seeds, the stateless policy even constructs the same arrangement that \method\ successfully verifies but dismantles it without testing it. Because a completed arrangement looks identical before and after testing, this difference shows that \method\ helps in remembering whether a candidate arrangement still needs to be verified, rather than in the preceding pick-and-place operation.

The remaining errors separate immediate verification from longer-term search. Six of the 14 stateless button presses miss the button, compared with none of the 35 \method\ presses, so motor precision further compounds the stateless failure rate. After a rejected arrangement, however, \method\ builds and tests a new order in only 10/28 rollouts; in other cases, it fails in rearranging the blocks or retests the rejected arrangement. These longer-search errors do not differ from the stateless policy. The differences in immediate verification and button execution both remain significant after Holm correction ($p{\leq}0.001$), but the broader pattern shows that memory supports the immediate transition from rearranging to verification more clearly than the full multi-attempt search.

Across all five tasks, the rollout analysis localises the performance gains shown in Section~\ref{sec:results} to the decisions that require past information in an episode. \method\ does not consistently change the early motor stages shared by both policies. Instead, it reduces the failure specific to each aliased decision: returning to the wrong pad, selecting the wrong block, declaring a swap complete too early, repeating an unsuccessful battery orientation, or dismantling an arrangement before testing it. The remaining failures separate into execution errors, such as stalling, missed presses, and dropped objects, and longer-term memory errors that arise when several updates must be retained. These behavioural patterns are consistent with the recurrent state retaining task-relevant episode information while preserving general robot control ability.
\FloatBarrier

\end{document}